%% file: main.tex
\documentclass{article}

\input{macros}

\usepackage[final, nonatbib]{neurips_2022}
\input{math_commands}

\usepackage[utf8]{inputenc} 
\usepackage[T1]{fontenc}    

\usepackage{booktabs,hyperref,url}       
\usepackage{amsfonts}       
\usepackage{nicefrac}       
\usepackage{microtype}      
\usepackage{xcolor}         

\title{Online Learning for Mixture of Multivariate Hawkes Processes}

\makeatletter
\newcommand{\printfnsymbol}[1]{%
  \textsuperscript{\@fnsymbol{#1}}%
}
\makeatother

\author{%
  Mohsen Ghassemi \and 
    \textbf{Niccol\`o Dalmasso} \and
    \textbf{Simran Lamba} \and 
    \textbf{Vamsi K. Potluru} \AND
    \textbf{Sameena Shah} \and
    \textbf{Tucker Balch} \and
    \textbf{Manuela Veloso} \AND
{\rm J.P. Morgan AI Research}\\[2pt]
$\{$\texttt{mohsen.ghassemi}, \texttt{niccolo.dalmasso}, \texttt{vamsi.k.potluru}$\}$\texttt{@jpmchase.com}
}

\begin{document}

\maketitle

\begin{abstract}
Online learning of Hawkes processes has received increasing attention in the last couple of years especially for modeling a network of actors. However, these works typically either model the rich interaction between the events or the latent cluster of the actors or the network structure between the actors. We propose to model the latent structure of the network of actors as well as their rich interaction across events for real-world settings of medical and financial applications.
Experimental results on both synthetic and real-world data showcase the efficacy of our approach.
\end{abstract}

\section{Introduction}

\input{Intro.tex}

\section{Problem Formulation}\label{sec:prob_def}

Suppose we observe a set of $N$ sequences $\mS=\{\vs^n\}_{n=1}^N$ in an online manner such that $\vs^n=\{e_i\triangleq (p_i,t_i)\}_{i=1}^{M_n}$ is the set of events (with time-stamp $t_i$ and event types $p_i\in \sP=\{1,\cdots, P\}$) observed so far. Each sequence corresponds to interactions (events) of a customer (actor node) with a product (indicating the event type). We model these event sequences using a mixture of Hawkes processes model. Specifically, for type $p$ sequences belonging to the $k$-th community the intensity function at time $t$ is
\begin{align}\label{mixture-hawkes}
\lambda^k_p(t) = \mu^k_p + \sum_{i: t_i<t} f^k(t-t_i; \theta_{p_i p})   
\end{align}
where $\mu^k_p$ is the base intensity of the event type $p$ for actors in community $k$, and $f^k(t-t_i; \theta_{p_i,p})  $ is the impact function of event type $p_i$ on event type $p$ for actors in community $k$, parameterized by $\theta^k_{p^i,p}$. In this work, we assume the impact functions are exponentials in form of $$f^k(t-t_i; \theta_{p_i ,p})  = a_{p_i, p}^k \exp\big(-b_{p_i, p}^k (t-t_i)\big),$$
where $\theta^k_{p_ip}=(a^k_{p_ip}, b^k_{p_ip})$. Define $\bm \theta^k=(\theta^k_{p^ip^j})_{p_i,p_j\in[P]}$, $\bm \mu^k= (\mu^k_{p})_{p_i,p_j\in[P]}$, and $\bm \Theta=\{\bm \mu^k,\bm \theta^k\}_{k=1}^K$. Then, the probability of observing a sequence $\vs$ 
can be described as 
\begin{align}
    P(\vs;\bm \Theta) = \sum_{k=1}^K \pi_k~ \mathrm{HP} (\vs|\bm \mu^k,\bm \theta^k)
\end{align}
where $\mathrm{HP} (\vs|\bm \mu^k,\bm \theta^k)= \prod\limits_{i: t_i<T} \lambda_{p_i}^k(t_i) \exp\Big(-\sum_{p=1}^P\int_{0}^T \lambda_p^k(s) ds\Big)$ \cite{rasmussen2018lecture} and $\{\pi^k\}$ are the probabilities of the clusters.

\subsection{Discretized Loss Function}

We discretize the point process by partitioning time into intervals of length $\delta$ and evaluate the likelihood functions at the end of the intervals. Denote by  $x_{\tau,p}=N_{p, \tau\delta}-N_{p, (\tau-1)\delta}$ the number of events of type $p$ that occurred during the $\tau^{\mathrm{th}}$ time interval. We can then discretize the conditional likelihood function $\mathrm{HP}^{\delta} (\vs|\bm \mu^k,\bm \theta^k)$ as 
\begin{align*}
    \mathrm{HP}^{\delta} (\vs|\bm \mu^k,\bm \theta^k)= \prod\limits_{\tau=1}^{T/\delta}\prod\limits_{p=1}^{P} \big(\lambda_{ p}^k(\tau\delta)\big)^{x_{\tau,p}}\cdot \exp\big(-\sum_{\tau=1}^{T/\delta} \sum_{p=1}^P \delta   \lambda^k_{p}(\tau\delta) \big).
\end{align*}
Consequently, the discretized likelihood function can be written as 
\begin{align}
    P^{\delta}(\vs;\bm \Theta) = \sum_{k=1}^K \pi_k~ \prod\limits_{\tau=1}^{T/\delta}\prod\limits_{p=1}^{P}\Big[ (\lambda^k_{p}(\tau\delta)\big)^{x_{\tau,p}} \exp\big(-\delta   \lambda^k_{p}(\tau\delta)) \Big].
\end{align}

\section{Online Learning Framework}\label{sec:algo}

\input{Algorithm}

\input{experiments}

\section{Conclusions and Future Work}
We considered the setting where we model the rich interaction between a network of actors with multiple event types and derive an online framework, utilizing the EM framework, for clustering these actors as well as learning the Hawkes process parameters. A possible future direction is to apply our proposed framework \textbf{OMMHP} for modeling the behavior of financial customers in the pandemic settings as well as to patient visits in the medical domain.

\textbf{Disclaimer.} This paper was prepared for informational purposes by the Artificial Intelligence Research group of JPMorgan Chase \& Co and its affiliates (“J.P. Morgan”), and is not a product of the Research Department of J.P. Morgan.  J.P. Morgan makes no representation and warranty whatsoever and disclaims all liability, for the completeness, accuracy or reliability of the information contained herein.  This document is not intended as investment research or investment advice, or a recommendation, offer or solicitation for the purchase or sale of any security, financial instrument, financial product or service, or to be used in any way for evaluating the merits of participating in any transaction, and shall not constitute a solicitation under any jurisdiction or to any person, if such solicitation under such jurisdiction or to such person would be unlawful.


\bibliographystyle{abbrv}


\appendix

\section{Appendix: Network Formulation}
\input{Network_short}

\section{Appendix: The M-step of \textbf{OMMHP}}\label{appendix:mstep}

In this section we include the details of two approaches for updating the Hawkes parameters $(\bm \mu^k,\bm \theta^k)$ (the M-step of OMMHP framework) for completeness.

\begin{figure*}[!h]
    \centering
    \includegraphics[width=1.0\linewidth]{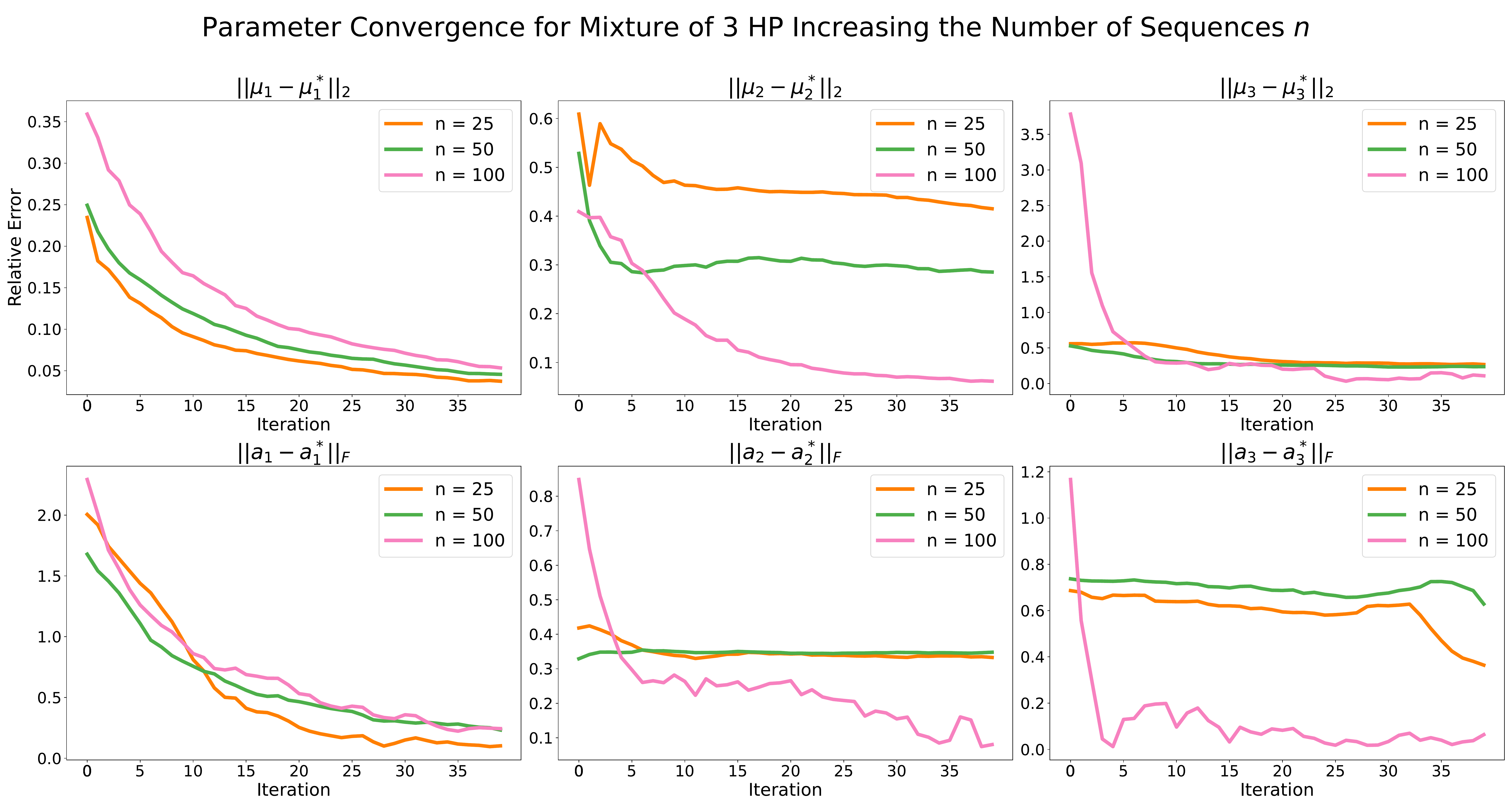}  
    \caption{Convergence plot for the mixtures of 3 Hawkes processes as a function of the generated number of sequences $n$. OMMHP relative errors are reported computing the $\ell_2$ norms (for baselines $\mu_i$) and Frobenius norms (for adjacency matrices $\bm a_i$) in solid lines. 
    For details see Section~\ref{sec:synth_data}.}
    \label{fig:convergence_function_realization_3HP}
\end{figure*}

\begin{figure*}[!h]
    \centering
    \includegraphics[width=0.465\linewidth]{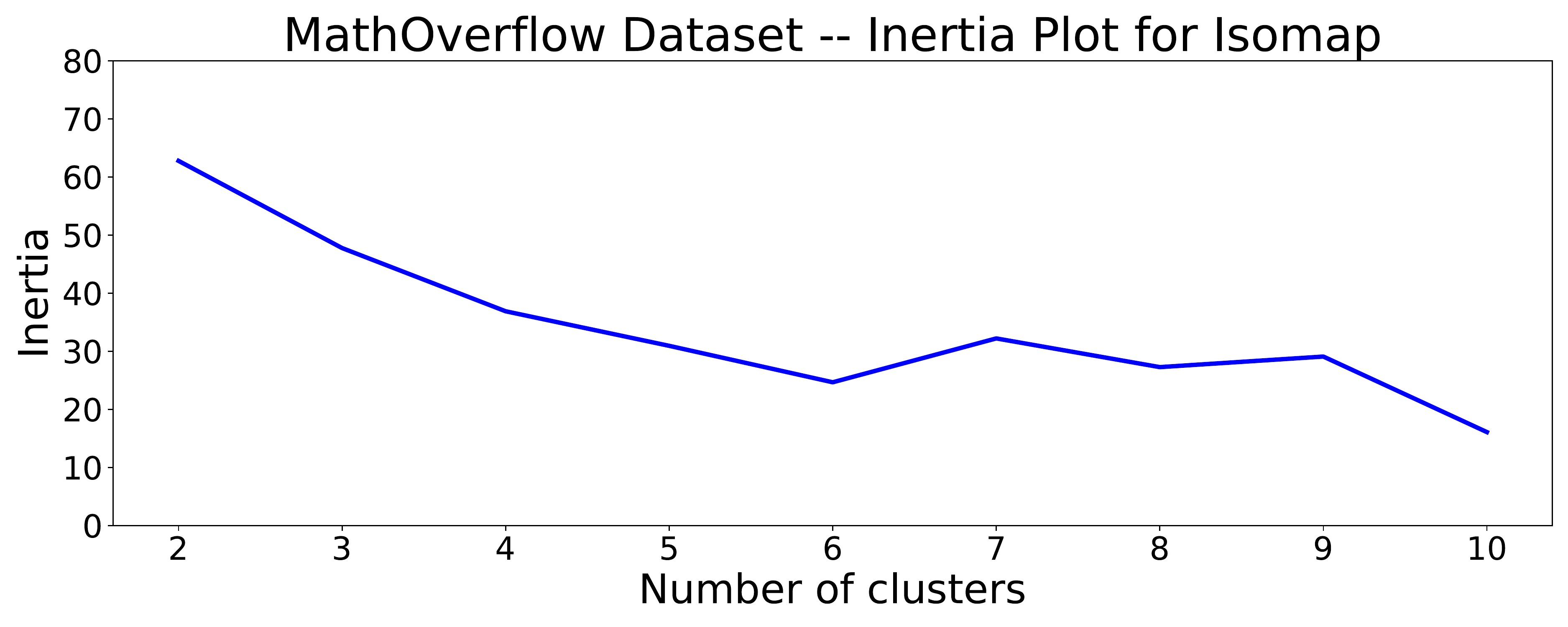}  
    \includegraphics[width=0.495\linewidth]{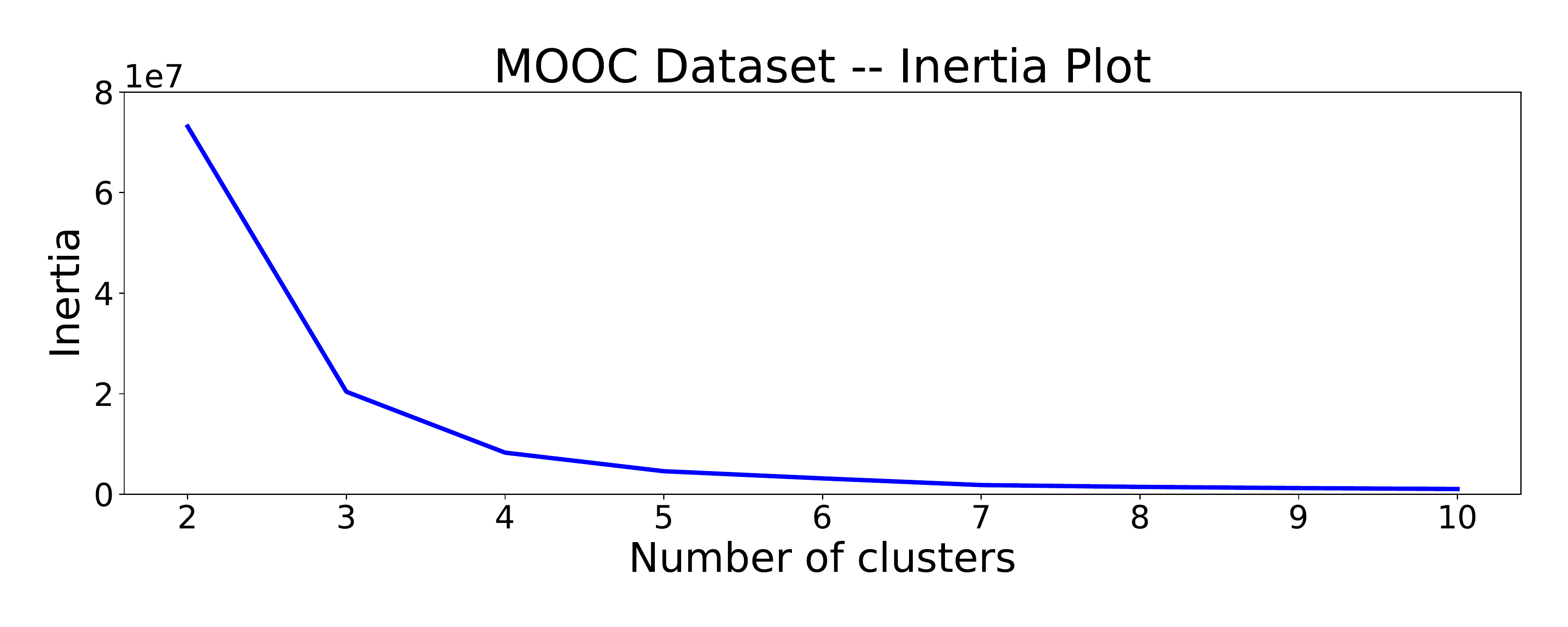}
    \caption{Inertia plot for the clustering of the MathOverflow dataset (left) and the MOOC (dataset). For the MathOverflow dataset,  we select 6 clusters based on the dramatic increase of the inertia after. For the MOOC dataset, we select 4 clusters using the elbow method. For details see Section~\ref{sec:real_data}.}
    \label{fig:elbow_plots}
\end{figure*}

\subsection{EM update}

One approach for estimating the parameters of a single Hawkes process (i.e. parameters $(\bm \mu^k,\bm \theta^k)$ in $\mathrm{HP} (\vs|\bm \mu^k,\bm \theta^k)$) is to apply an EM approach. We refer the readers to  \cite{zipkin2016hawkes} and \cite[Chapter 5]{morse2017persistent} for the details of this approach.

\subsection{Stochastic gradient update} \label{sec:sgd_updates}

While in our experiments we use the EM approach, due to its better performance, we include the gradient update on the parameters of the Hawkes processes in the stochastic gradient approach for completeness. 

The (stochastic) gradient update on the parameters of the Hawkes processes are as follows: 
\begin{align*}
     \mu^k_p(t) =  \mu^k_p({t-1}) + \eta_{t} \frac{\partial \E_{q^{t-1}(\mZ)}L_{t}^{\delta}(\bm \Theta, \mZ)}{\partial  \mu_p^k }
\end{align*}
\begin{align*}
    \theta^k_{p^i,p^j}(t) =  \theta^k_{p^i,p^j}(t-1) + \eta_{t} \frac{\partial \E_{q^{t-1}(\mZ)}L_{t}^{\delta}(\bm \Theta, \mZ)}{\partial  \theta^k_{p^j,p^j} }
\end{align*}
Specifically, the gradient with respect to $\mu_p^k$ and $\theta^k_{p_ip}=(a^k_{p_ip}, b^k_{p_ip})$ can be calculated as follows: 
%
%
\begin{align*}
     \frac{\partial \E_{q^{t-1}(\mZ)}L_{t}^{\delta}(\bm \Theta, \mZ)}{\partial \mu_{p}^k }=\sum_{n=1}^N \alpha_{nk} \big[  \frac{x^{t}_{n,p}}{\lambda^k_{n,p}(t\delta)} -\delta  \big] 
\end{align*}
and the gradient with respect to $\theta^k_{p_ip}=(a^k_{p_ip}, b^k_{p_ip})$ is calculated as follows: 
%
\begin{align*}
     &\frac{\partial \E_{q^{t-1}(\mZ)}L_{t}^{\delta}(\bm \Theta, \mZ)}{\partial a^k_{p_ip_j} } =  \sum_{n=1}^N \alpha_{nk} \big[  \frac{x^{t}_{n,p_j} }{\lambda^k_{n,p_j}(t\delta)} -\delta  \big] \sum_{I_n\in \sP^t_{p_i}}\exp\big(-b_{p_{I_n}, p_j}^k (t\delta-t_{I_n})\big),
\end{align*}
\begin{align*}
     &\frac{\partial \E_{q^{t-1}(\mZ)}L_{t}^{\delta}(\bm \Theta, \mZ)}{\partial b^k_{p_ip_j} }=  \sum_{n=1}^N\Big[ \alpha_{nk} \big[  \frac{x^{t}_{n,p_j} }{\lambda^k_{n,p_j}(t\delta)} -\delta  \big]\cdot\sum_{I_n\in \sP^t_{p_i}} a^k_{p_{I_n}p_j} (t_I-t\delta)\exp\big(-b_{p_{I_n}, p_j}^k (t\delta-t_{I_n})\big)\Big] 
\end{align*}
where $\sP^{t,n}_{p_i}=\{e_{I_n}=(t_{I_n},p_{I_n})~|~t_{I_n}\leq t\delta,~ p_{I_n}=p_i\}$.

Note that while the vanilla stochastic gradient method does not perform well for the maximum likelihood objective of Hawkes Processes, applying variance reduction methods such as SVRG \cite{johnson2013accelerating} may improve the performance. We leave exploring this direction to future work.

\section{Appendix: Experiments} \label{appendixc: experiments}

In this section we include the figures and experiments which were not included in the main paper for easiness of read. Figure~\ref{fig:convergence_function_realization_3HP} includes the parameter convergence behavior for a mixture of 3 Hawkes processes in the experiment settings presented in Section~\ref{sec:synth_data}. Figure~\ref{fig:elbow_plots} reports the inertia plot for the clustering of the MathOverflow and MOOC datasets of Section~\ref{sec:real_data}, which are used for guidance on selecting the number of clusters.

\begin{figure*}[!h]
    \centering
    \includegraphics[width=1.0\linewidth]{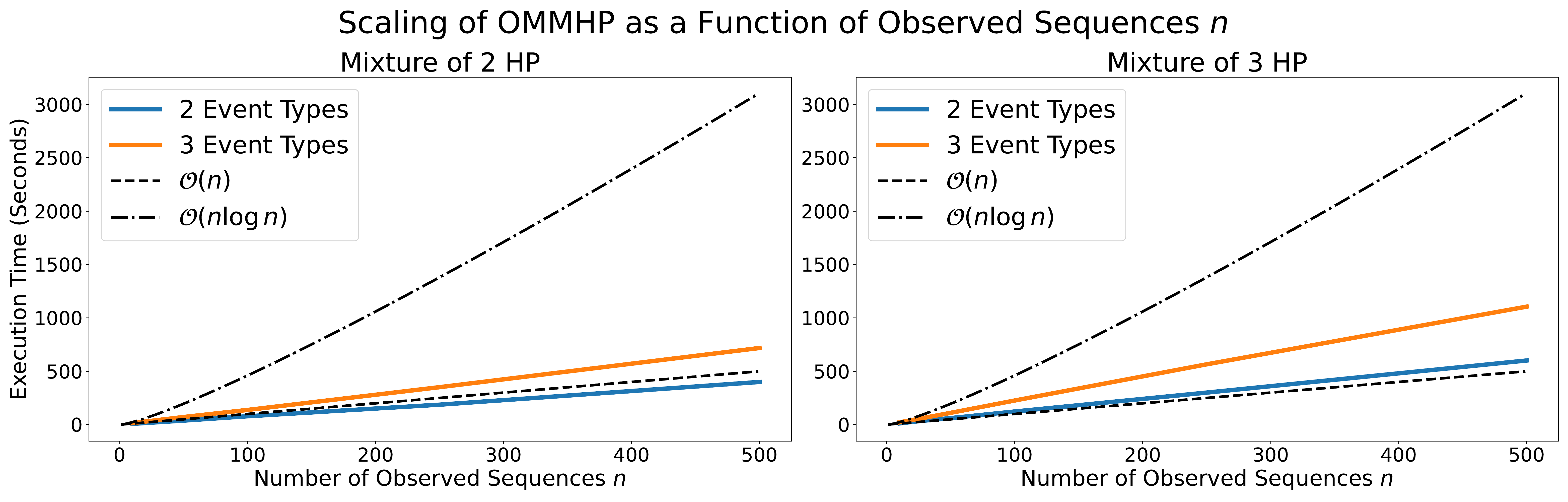}  
    \caption{Execution time for \textbf{OMMHP} for a mixture of two (left) and three (right) Hawkes processes as a function of the observed event sequences $n$. The $\mathcal{O}(n)$ and $\mathcal{O}(n \log n)$ are included for comparison purposes.  For details see Section~\ref{sec:synth_data}}
    \label{fig:scaling_plot_ommhp}
\end{figure*}

\section{Appendix: Scalability Analysis}

Figure~\ref{fig:scaling_plot_ommhp} shows the execution time for \textbf{OMMHP} for a mixture of two (left) and three (right) Hawkes processes, using the synthetic data setting of Section~\ref{sec:synth_data}. We can see that the algorithm scales roughly linearly with the number of observed event sequences $n$. Adding more event types and/or mixture components seem also to increase the computation time in a linear fashion. More specifically, adding either an extra event type or mixture seem to increase the running time by between $50\%$ and $100\%$, as reported by Table~\ref{tab:scaling} in which the running time normalized by number of sequences is reported.

\begin{table}[!h]
\centering
\begin{tabular}{|c|c|c|}
\multicolumn{3}{c}{\textit{OMMHP Runtime (Normalized, seconds)}} \\
\hline
& 2 Processes & 3 Processes \\
\hline
 2 Events & 0.77 $\pm$ 0.02 & 1.226 $\pm$ 0.05 \\
 3 Events & 1.418 $\pm$ 0.02 & 2.213 $\pm$ 0.04 \\
\hline
\end{tabular}
\vspace{0.5\baselineskip}
\caption{Average and standard deviation for OMMHP runtime in seconds, normalized by the number of observed sequences $n$. Both adding an process to the mixture or an event type seem to increase the execution time linearly, with adding the event type being the more computationally intensive.} \label{tab:scaling}
\end{table}

\end{document}

%% file: macros.tex
\usepackage{mathtools}
\usepackage{amssymb,latexsym,amsmath, mathtools,bm,amsthm}
\usepackage{color}
\usepackage[dvipsnames]{xcolor}
\usepackage{amssymb}
\usepackage{bbm,mathrsfs}
\usepackage{multicol,multirow}
\usepackage{setspace}

\def\mb{\mathbb}

\newcommand{\E}{\mb{E}}

\newcommand\redout{\bgroup\markoverwith{\textcolor{red}{\rule[.5ex]{2pt}{0.4pt}}}\ULon}

\usepackage{pifont}
\newcommand{\cmark}{\ding{51}}%
\newcommand{\xmark}{\ding{55}}%

%% file: math_commands.tex
\usepackage{amsmath,amsfonts,bm}
\usepackage{algorithmic,algorithm}
\usepackage{mathtools}
\usepackage{latexsym,amsmath, mathtools}
\usepackage{color}
\usepackage{bbm,mathrsfs}

\usepackage{pifont}

\def\eqref#1{equation~\ref{#1}}

\def\1{\bm{1}}

\def\vs{{\bm{s}}}

\def\vz{{\bm{z}}}

\def\mS{{\bm{S}}}

\def\mZ{{\bm{Z}}}

\DeclareMathAlphabet{\mathsfit}{\encodingdefault}{\sfdefault}{m}{sl}
\SetMathAlphabet{\mathsfit}{bold}{\encodingdefault}{\sfdefault}{bx}{n}

\def\sP{{\mathbb{P}}}

%% file: OMMHP Arxiv/intro.tex
In many applications we need to deal with large number of sequences of events that occur asynchronously and at irregular intervals. Examples of such sequential data include interactions of customers with a bank (customers journeys), customers purchases at a grocery/online store, visits of patients to a hospital, and spread of viral diseases such as COVID-19~\cite{yan2019recent, mei2017neural,zuo2020transformer,browning2021simple, garetto2021time}. Each event sequence can consist of multiple events of different types. 
Predicting the time and the type of the next event for each sequence, finding the possible latent cluster of actors~\cite{xu2017dirchlet}, inferring causal relations between events~\cite{xu2016causality}, and deliberately influencing the behaviors of the actors~\cite{farajtabar2014shaping}~are some interesting applications when dealing with sequential data.

Due to irregular and asynchronous nature of sequential events data, the standard time-series approaches are not able to capture the rich information that exist in occurrence times of such data. Instead, point processes are commonly used to learn the distribution of sequential event data. Specifically, multivariate Hawkes processes (MHP) have received considerable attention in the recent years due to their ability to model triggering (or inhibiting) effect of past events of different type on future events~\cite{rasmussen2018lecture}. Many works in the literature focus on employing non-parametric as well as neural network based approaches to design impact functions that are able to model complex dependencies of events in point processes~\cite{mei2017neural,zuo2020transformer,zhang2019self}. Despite the outstanding results, a limitation of the aforementioned works is that they learn a single dependency pattern for all the sequences of events. To address this shortcoming, \cite{xu2017dirchlet} propose a \textit{mixture of multivariate Hawkes processes} (MMHP) which  allow for modeling event
sequences where there exist multiple impact
patterns across the
sequences. The mixture model allows for identifying latent clustering structures across event sequences. 

In this work, we propose \textit{online learning for mixture of multivariate Hawkes processes} (\textbf{OMMHP}). The online learning framework addresses scalability issues of batch models by reducing the per iteration computational complexity. Moreover, in many applications such as banking, shopping, and health care we deal with streaming sequences of events. It is often the case that the behaviors of the actors, i.e. the dependencies between events, change over time, possibly abruptly. Similarly, the cluster structures can change over time, with people moving from one cluster to another (e.g. due to a change in employment status) or with new clusters emerging and old clusters disappearing. The drastic changes that many communities are going through due to the recent COVID-19 pandemic emphasizes the dynamic nature of the underlying models of sequential data in many applications \cite{garetto2021time,chiang2020hawkes}. Online learning of MMHP allows to identify and model this dynamic nature without the need for the costly process of training models from scratch. To the best of our knowledge, this paper presents the first online MMHP framework that takes into account the interactions and dependencies between different types of events.



 \subsection{Related Work}
 

\paragraph{Clustering of Point Processes}
While many of the existing work on clustering point processes focus on clustering subsequences of events within the sequences in order to find common patterns \cite{Du2015dirichlet,turkmen2020clustering}, some recent art have addressed the challenging problem of clustering the sequences themselves. Most notably, \cite{xu2017dirchlet} propose a probabilistic mixture model that aims at modeling sequences that are generated from a mixture of self-triggering Hawkes processes. Along similar lines, \cite{sharma2019generative} propose a generative model that allows for modeling mixtures of more complex point processes through the use of recurrent models such as RNNs and LSTMs.
 
 \paragraph{Online Learning of Point Processes}
 
\cite{hall2016tracking} propose a parametric online framework for learning Hawkes processes. In contrast, \cite{yang2017online} propose a non-parametric approach that allows for more complex interaction functions. Most recently, \cite{fang2020online} proposed an online community detection algorithm based on sequences of interactions among nodes in a network. We propose a general model which encompasses the setting in \cite{fang2020online} by considering the multivariate case where we allow for multiple event types (types of interaction) to occur between the nodes.
\begin{table}[t!]
\caption{Prior art on online learning of point processes}
\label{table:network}
\centering
\begin{tabular}{|c|c|c|c|}
\hline
Paper & sequence clustering & event interactions & network \\
\hline
\cite{hall2016tracking} & \xmark  &  \xmark & \cmark\\
\hline
\cite{yang2017online} & \xmark  &  \cmark  & \xmark\\
\hline
\cite{fang2020online} & \cmark  &  \xmark  & \cmark\\
\hline
This work & \cmark  &  \cmark  & \cmark\\
\hline
\end{tabular}
\end{table}
Table~\ref{table:network} contains a comparison of our model versus the existing models in the literature. For ease of exposition, we present an online framework for learning the mixture of Hawkes processes with event interactions and defer the full version of incorporating the network structure to the appendix.  


%% file: Algorithm.tex
It is difficult to directly work with the log-likelihood $$\log \big(\sum_k \pi^k \mathrm{HP}^{\delta}(\vs|\bm \mu^k,\bm \theta^k) \big)$$ due to summation inside the logarithm. 
Instead, we work with the evidence lower bound (ELBO) defined as 
\begin{align}
    \mathrm{ELBO}(\bm \Theta)= \E_{q(\mZ)}[L^{\delta}(\bm \Theta,\mZ)]- \E_{q(\mZ)}[\log q(\mZ)]
\end{align}
for a properly chosen distribution $q(\mZ)$, where $L^{\delta}(\bm \Theta, \mZ)$ is the complete log-likelihood defined as 
%
\begin{align}
    L^{\delta}(\bm \Theta, \mZ)\triangleq& \sum_{n=1}^N\sum_{k=1}^K z_{nk}\Big[ \log \pi^{k} + \log \mathrm{HP}^{\delta}(\vs_n|\bm \mu^k,\bm \theta^k) \big)\Big]\nonumber\\
    %
    %
    = & \sum\limits_{\tau=1}^{T/\delta}\sum_{n=1}^N \sum\limits_{p=1}^{P}\sum_{k=1}^K z_{nk}\big[ x^{\tau}_{n,p} \log (\lambda^k_{n,p}(\tau\delta)\big) -\delta   \lambda^k_{n,p}(\tau\delta) \big] + \sum_{n=1}^N \sum_{k=1}^K z_{nk} \log \pi^{k} \nonumber \\
    =&  \sum\limits_{\tau=1}^{T/\delta} L_{\tau}^{\delta}(\bm \Theta, \mZ)  + \sum_{n=1}^N \sum_{k=1}^K z_{nk} \log \pi^{k} 
\end{align}
is the complete log-likelihood function. The additive nature of $L^{\delta}(\bm \Theta, \mZ)$ allows for adopting online algorithms to maximize the ELBO function. We assume that $q(\mZ)$ takes the simple form $q(\mZ)=\prod_{n=1}^N q(\vz_n)$ where $q(\vz_n)\sim \mathrm{multinom}(\bm\alpha_n)$ and $\vz_n\in [K]$. Therefore,
\begin{align}
    \mathrm{ELBO}(\bm \Theta)
    &= \sum\limits_{\tau=1}^{T/\delta} \E_{q(\mZ)}L_{\tau}^{\delta}(\bm \Theta)  + \sum_{n=1}^N \sum_{k=1}^K \alpha_{nk} \log \pi^{k}/\alpha_{nk} 
\end{align}
with
\begin{align*}
 \E_{q(\mZ)}L_{\tau}^{\delta}(\bm \Theta, \mZ)  =  \sum_{n=1}^N \sum\limits_{p=1}^{P} \sum_{k=1}^K \alpha_{nk} \big[ x^{\tau}_{n,p} \log (\lambda^k_{n,p}(\tau\delta)\big) -\delta   \lambda^k_{n,p}(\tau\delta) \big] .
 \end{align*}
This allows us to employ the online variational inference framework for mixture models employed by \cite{sato2001online}, \cite{hoffman2010online}, and \cite{fang2020online}. The $t$-th iteration of  OMMHP consists of the following two steps:

\paragraph{E-Step (responsibilities update):} Due to the independence assumption on cluster assignments ($q(\mZ)=\prod_{n=1}^N q(\vz_n)$), we can adopt a coordinate descent procedure for updating the responsibilities, i.e., we alternatingly maximize the ELBO function with respect to $\bm \alpha_n$:
\begin{align}
    \alpha_{nk}^t &\propto \pi^k(t-1) \underbrace{\exp\left(\sum_{\tau=1}^t\sum_{p=1}^P \big[ {x^{\tau}_{n,p}} \log (\lambda^k_{n,p}(\tau\delta)\big) -\delta   \lambda^k_{n,p}(\tau\delta) \big]\right)}_{R^{t}(k,n)}\nonumber
    %
\end{align}
where $$R^{t}(k,n) = \exp \left(\sum\nolimits_{p=1}^P \big[ x^{t}_{n,p} \log (\lambda^k_{n,p}(t\delta)\big) -\delta \lambda^k_{n,p}(t\delta) \big]\right)\cdot R^{t-1}(k,n)$$ can be computed recursively and $\pi^k(t-1)$ is the estimate of $\pi^k$ after observing the $(t-1)$-th interval (i.e. at the beginning of the $t$-th interval). It follows trivially that $\alpha_{nk}^t=\frac{\pi^k(t-1) R^{t}(k,n)}{\sum_{k=1}^K \pi^k(t-1) R^{t}(k,n)}$ and $\pi^k(t)=\sum_{n=1}^N \alpha_{nk}^t /N$.

\paragraph{M-Step (HP parameters update):} 

The M-step of \textbf{OMMHP} framework involves updating parameters $\mu^k_p$ and $\theta^k_{p^i,p^j}$. This can be done by a variety of methods including expectation-minimization (EM) as proposed in \cite[Chapter 5]{morse2017persistent}, or using stochastic gradient updates. More details on the M-step can be found in Appendix B.  
The pseudo-code for \textbf{OMMHP} 
is presented in Algorithm \ref{algo:ommhp}. 

 \begin{algorithm}[t]
\caption{Online mixture of multivariate Hawkes Processes (\textbf{OMMHP})} \label{algo:ommhp}
\begin{algorithmic}[1]
\REQUIRE $K$, $P$, $T$, $\mS$ , $\delta$
\STATE Initialize $\{R^0(k,n)\}$,  $\{\bm\alpha_n(0)\}$, $\{\bm\mu^k(0)\}$, $\{\bm\theta^k(0)\}$
\FOR{$t=1$ to $T/\delta$}
\STATE Observe new events in $(\delta(t-1),\delta t]$ for all sequences
\STATE Set the step size $\eta_{t}$ 
%
%
\STATE E-step:
\FOR{$n=1$ to $N$}
\FOR{$k=1$ to $K$}
\FOR{$p=1$ to $P$}
\STATE $\lambda^k_{n,p}(t\delta) = \mu^k_{p}(t-1) + \sum\limits_{i_n: t_{i_n}<t} f^k(t\delta-t_{i_n}; \theta_{p_{i_n} p}(t-1))$
\ENDFOR
\STATE $R^{t}(k,n) = \exp(\sum\limits_{p=1}^P \big[ x^{t}_{n,p} \log (\lambda^k_{n,p}(t\delta)\big) -\delta \lambda^k_{n,p}(t\delta) \big])\cdot R^{t-1}(k,n)$
\ENDFOR
\FOR{$k=1$ to $K$}
\STATE $\alpha_{nk}^t=\frac{\pi^k(t-1) R^{t}(k,n)}{\sum_{k=1}^K \pi^k(t-1) R^{t}(k,n)}$
\ENDFOR
\ENDFOR
\FOR{$k=1$ to $K$}
\STATE $\pi^k(t)=\sum_{n=1}^N \alpha_{nk}^t /N$
\ENDFOR
\STATE M-step: 
\STATE Use estimation algorithm for $\mu_p^k$ and $\theta^k_{p^i, p^j}$, such as EM \cite{zipkin2016hawkes} or stochastic gradient descent (see Appendix~\ref{sec:sgd_updates})

%

%
\ENDFOR
\end{algorithmic}
\end{algorithm}

%% file: experiments.tex
\section{Experiments}  
We validate the effectiveness of \textbf{OMMHP} in learning the underlying parameters of a mixture of Hawkes processes on both synthetic and real-world data.

\subsection{Synthetic Data} \label{sec:synth_data}
We generate two sets of synthetic datasets mixing the following Hawkes processes:

\begin{align*}
    \mathbf{x} &\sim HP\left(\bm \mu_1 = [0.3, 0.2], \bm a_1 =
\begin{bmatrix}
     0.2 & 0.1 \\ 0.1 & 0.2
\end{bmatrix} \right) \\
\mathbf{y} &\sim HP\left(\bm \mu_2 = [2.8, 1.6], \bm a_2 = 
\begin{bmatrix}0.6 & 0.2 \\ 0.2 & 0.6
\end{bmatrix} \right) \\
\mathbf{z} &\sim HP\left(\bm \mu_3 = [1.4, 0.8], \bm a_3 =
\begin{bmatrix}
     0.4 & 0.0 \\ 0.0 & 0.6
\end{bmatrix} \right)
\end{align*}

In the first dataset we generate $n=10$ sequences from the first two processes, so $\mathcal{D}_1 = \left \{ (\mathbf{x_i})_{i=1}^{n}, (\mathbf{y_j})_{j=1}^{n}  \right \}$, while in the second dataset we generate $n=10$ sequences from all processes, so $\mathcal{D}_2 = \left \{ (\mathbf{x_i})_{i=1}^{n}, (\mathbf{y_j})_{j=1}^{n}, (\mathbf{z_k})_{k=1}^{n}  \right \}$. In both datasets we set $T=1000$ and the decay rate to $u = 3.1$.

\begin{figure*}[!h]
    \centering
    \includegraphics[width=0.53\linewidth]{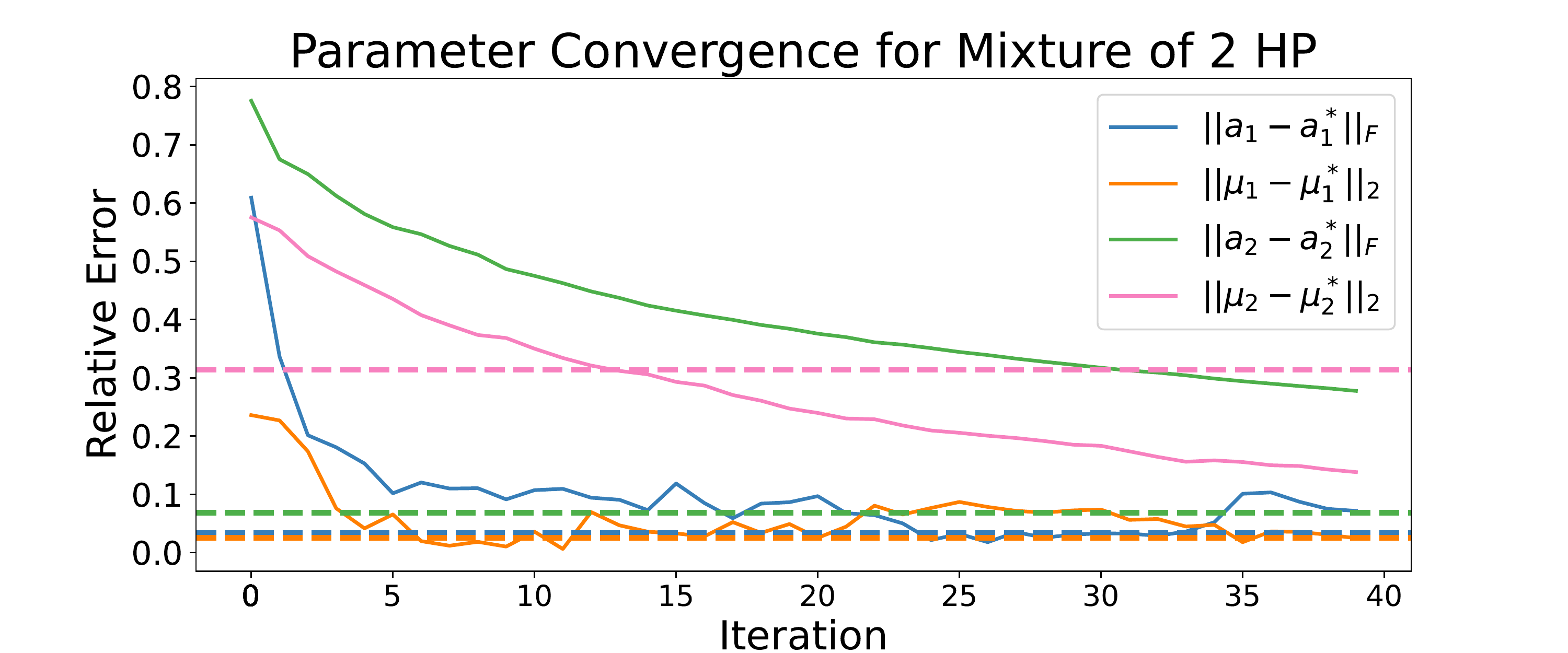}  
    \hspace{-1cm}
    \includegraphics[width=0.53\linewidth]{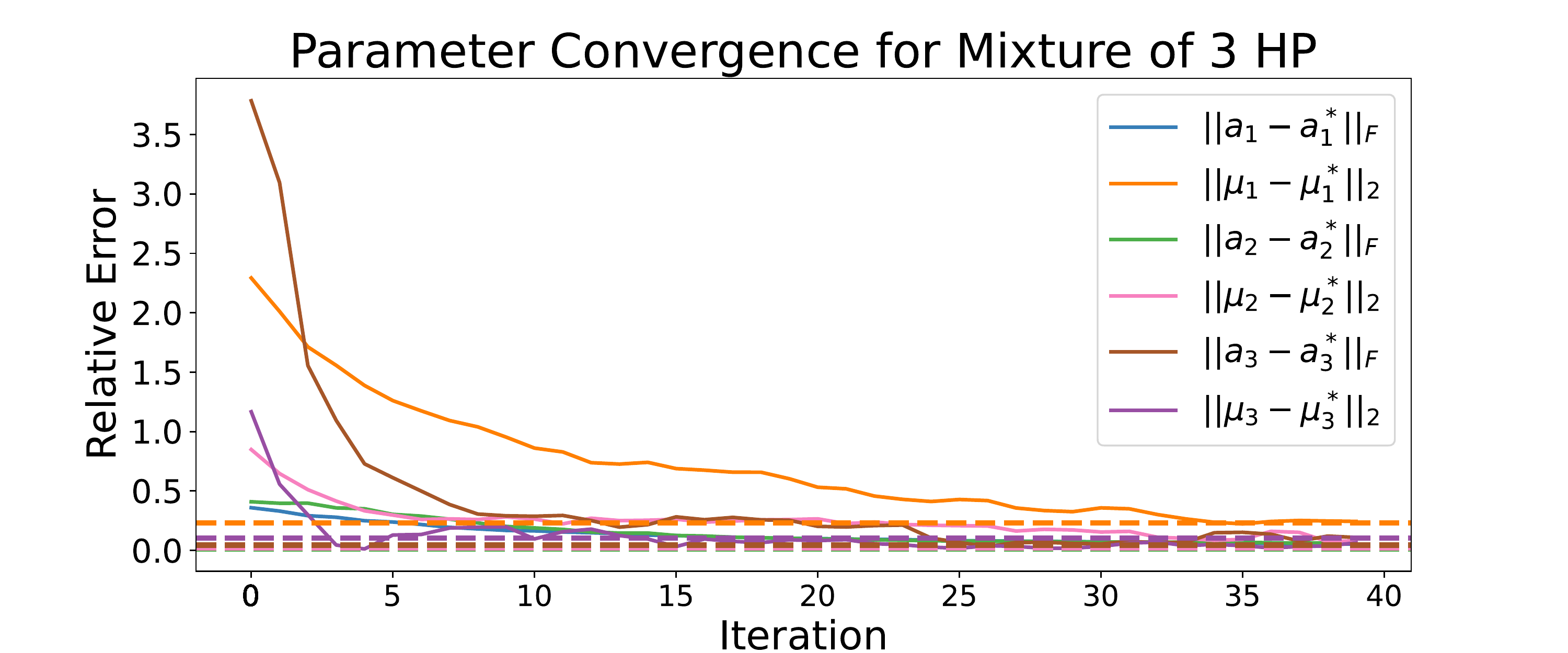}
    \caption{Convergence plot for the mixtures of 2 (left) and 3 (right) Hawkes processes with $n=10$ observed sequences. OMMHP relative errors are reported computing the $\ell_2$ norms (for baselines $\mu_i$) and Frobenius norms (for adjacency matrices $\bm a_i$) in solid lines. Dashed lines indicate the relative error achieved by accelerated gradient descent when using true cluster labels. OMMHP correctly recovers the correct cluster assignment and achieves relative errors which are comparable to off-the-shelf algorithms that have access to the true cluster labels. For details see Section~\ref{sec:synth_data}.}
    \label{fig:convergence}
\end{figure*}

\begin{figure*}[!h]
    \centering
    \includegraphics[width=1.0\linewidth]{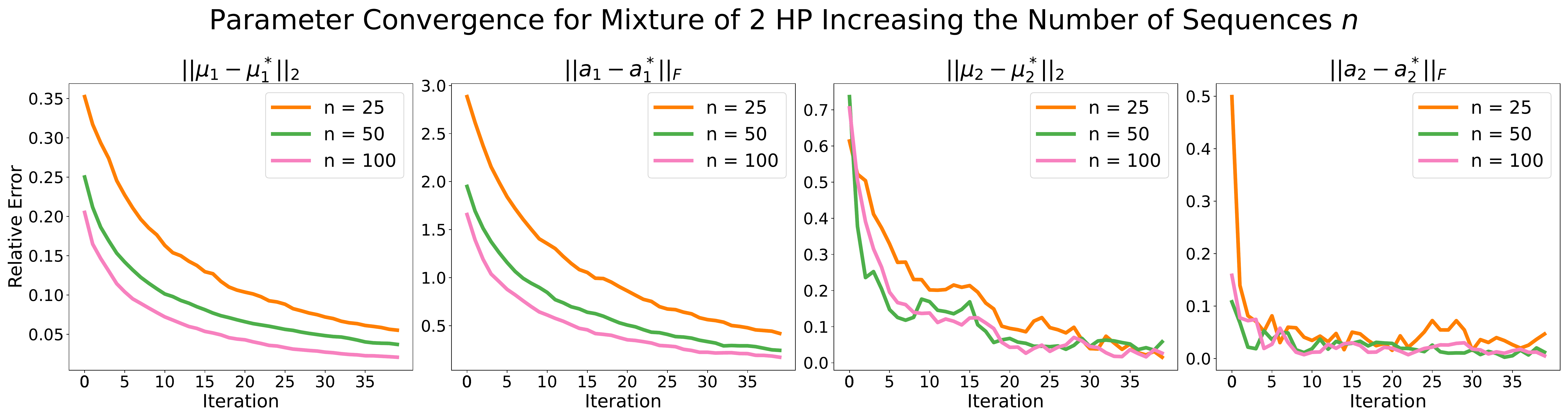}  
    \caption{Convergence plot for the mixtures of 2 Hawkes processes as a function of the generated number of sequences $n$. OMMHP relative errors are reported computing the $\ell_2$ norms (for baselines $\mu_i$) and Frobenius norms (for adjacency matrices $\bm a_i$) in solid lines. As expected, a higher number of generated sequences $n$ leads to a faster convergence, although with different behaviors between the two processes in the mixture. For details see Section~\ref{sec:synth_data}.}
    \label{fig:convergence_function_realization_2HP}
\end{figure*}

We apply \textbf{OMMHP}  on both these dataset to recover the mixture identities of the sequences and learn the parameters of the Hawkes processes. 
We chose $\delta=25$ and update the parameters using a $\eta_t=\frac{1}{\sqrt{t+1}}$ decay, a standard decay rate commonly used in online learning literature \cite{hall2016tracking}. In both cases the correct cluster assignments is learned by \textbf{OMMHP}. Moreover, Figure~\ref{fig:convergence} shows the convergence of the learned parameters to the true parameters as a function of the iteration. We compute the relative difference in $\ell_2$ norm for vectors and Frobenius norms for matrices for the learned and true parameter values. As baseline, we provide the relative error in estimating each cluster parameters using accelerated gradient descent (using the \texttt{tick} package by \cite{bacry2017tick}, shown in dashed lines in Figure~\ref{fig:convergence}). \textbf{OMMHP} achieves relative errors which are comparable to off-the-shelf algorithms that have access to the true cluster labels. In addition, we investigate the convergence behavior as a function of the number of generated sequences $n$ for each process. Figure~\ref{fig:convergence_function_realization_2HP} shows that for the mixture of two Hawkes processes the parameter convergence is generally faster if a higher number of sequences from each process is observed, which is the expected behavior. However, a higher number of sequences $n$ improves the estimation error more significantly for the first process in the mixture, which indicates the first process is the hardest to estimate among the two. A similar takeaway applies to the mixture of three Hawkes processes case, which we include in Figure~\ref{fig:convergence_function_realization_3HP} in Appendix~\ref{appendixc: experiments}. Finally, we also include an empirical study showing the estimation process scales roughly linearly in the number of generated sequences $n$, event types, and mixture components; see Figure~\ref{fig:scaling_plot_ommhp} and Table~\ref{tab:scaling} in Appendix~\ref{appendixc: experiments} for details.

\subsection{Real Data} \label{sec:real_data}

To showcase \textbf{OMMHP}'s performance on real-world data we consider two real-world datasets: (1) the \textit{MathOverflow} dataset \cite{leskovec2014snap}, which tracks the interaction between users on a question-answering forum between the years of 2009 and 2016 and (2) the \textit{MOOC} dataset \cite{feng2019dropout}, which tracks the actions of students of a massive online open course held virtually by a Chinese university. Throughout this section we use the adjusted Rand index (ARI, \cite{rand1971objective}) as clustering performance metric. If the clustering performance is roughly equivalent to random clustering, the ARI will be equal to zero, while a better-than-chance performance is reflected by a positive ARI. For both experiments we fit OMMHP using Algorithm~\ref{algo:ommhp} with $\eta_t=\frac{1}{\sqrt{t+1}}$.

\subsubsection{MathOverflow Dataset}

\begin{figure}[!h]
    \centering
    \includegraphics[width=0.75\linewidth]{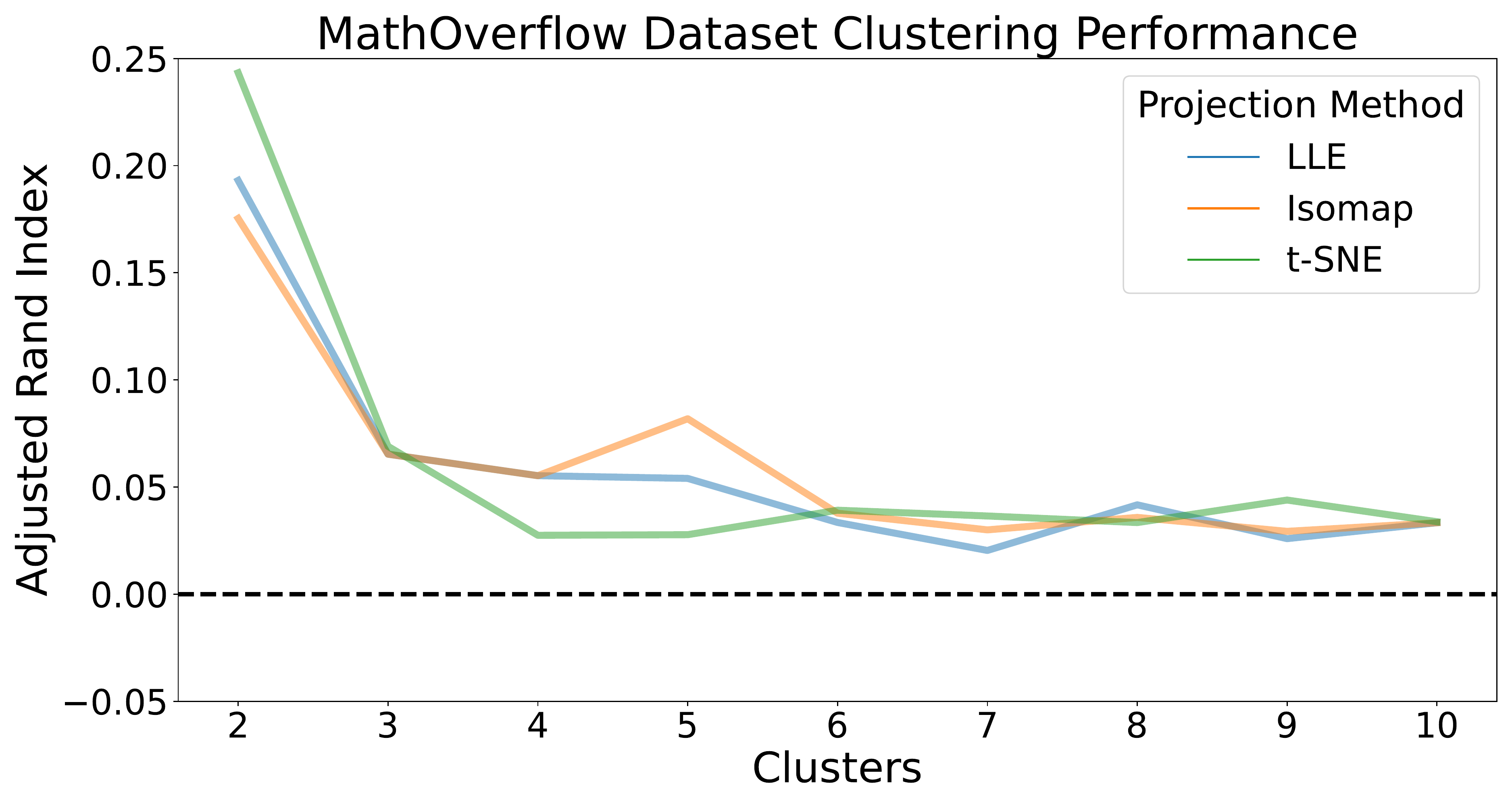}  
    \caption{Clustering performance (indicated by ARI) across various clusters and projection methods for the MathOverflow dataset. 
    For details see Section~\ref{sec:real_data}.}
    \label{fig:mathoverflow_clustering_performance}
\end{figure}

This dataset is structured so that each interaction is recorded as a pair of users (sender and receiver) and a timestamp for when the interaction occurred. As there are no details on the interaction type, we grouped all interactions together as a single even type. We select the interactions coming from the 100 most active users, selected as the union of the set of the 50 most active senders and 50 users who have received the most interactions in the dataset, for a total of $\sim 170,000$ interactions. We first compute a series of activity-related features, such as the total years of activity, total interactions and average monthly interactions, project such features using the Isomap algorithm \cite{tenenbaum2000global}\footnote{We found that clustering in the original feature space did not lead to any meaningful clustering behavior.} and cluster the users using the \texttt{k-means++} clustering algorithm. We select $6$ as the final number of clusters using the elbow method \cite{marutho2018determination,bock2008inertiakmeans}.
(see Figure~\ref{fig:elbow_plots}, left, in Appendix~\ref{appendixc: experiments}). The results is three large cluster (more than 20 users), one medium clusters (with 12) and a series of small size cluster (with less than 8 users each). OMMHP clustering achieves a better-than-chance final ARI of 0.1 on such clustering. To demonstrate the robustness of our approach, Figure\ref{fig:mathoverflow_clustering_performance} shows that OMMHP achieves better-than-chance clustering performance across various number of clusters (between 2 and 10) and projection algorithms (original data space, Isomap, t-SNE \cite{vandermaaten2008tsne} and locally linear embedding \cite{saul2000introduction}).

\begin{table*}[!h]
\centering
\begin{tabular}{|c|c|c|c|c|c|c|c|c|c|}
\multicolumn{10}{c}{\textit{MOOC Clustering Performance}} \\
\hline
 & \multicolumn{9}{|c|}{\textbf{Clusters}} \\
 \hline
 & 2 & 3 & \textbf{\textcolor{red}{4}} & 5 & 6 & 7 & 8 & 9 & 10 \\
\hline
\textbf{Adjusted Rand Index} & -0.047 & 0.025 & \textbf{\textcolor{red}{0.201}}  & 0.150 & 0.00 & 0.129& 0.107& 0.064& 0.098 \\
\hline
\end{tabular}
\vspace{0.5\baselineskip}
\caption{Clustering performance (indicated by ARI) across various cluster numbers for the MOOC dataset. Using OMMHP for detecting clusters mostly results in a better-than-chance clustering performance, indicated by an ARI larger than zero. The performance for the number of clusters chosen via the ``elbow method''is highlighted in red. For details see Section~\ref{sec:real_data}.} \label{tab:mooc_performance}
\end{table*}

\subsubsection{MOOC Dataset}

The MOOC dataset tracks the behavior of around $150,000$ students enrolled in 250 online open courses held by a Chinese university. All student actions, such as play a video, stop a video or click on a problem, are recorded in the dataset along with the timestamp of the action. In order to reduce the dimensionality of the dataset, we first select the most common 9 actions and group all other actions into the same category, hence creating 10 event types. Some students in the dataset perform almost the same action thousand of times, we filter such students. We then select the top 25 students across all courses, leaving a total of $\sim 365,000$ actions. Similarly to what performed above, we cluster the students based on the actions performed during their experience with the online class using the \texttt{k-means++} clustering algorithm; we select $4$ as the final number of clusters, since the inertia stops decreasing significantly when using more clusters (Figure~\ref{fig:elbow_plots}, right, for the inertia plot). The four clusters are composed of three balanced clusters with 8 students each, and cluster with a single student, for which using OMMHP for clustering achieves a better-than-chance ARI of $0.2$. Similarly for what shown above for the MathOverflow dataset, Table~\ref{tab:mooc_performance} reports the ARI for different cluster numbers, showing how using OMMHP for clustering usually achieves better-than-chance clustering results.

%% file: Network_short.tex
Consider an online stream of events of the form $s=\{e,t)\}$ with timestamp $t$ and event $e$. 
We assume that the event data corresponds to an interaction between two nodes on a network $A$. 
Mathematically, we can enumerate the set of events for a network as $\mathbf{E} = \{(i,j,p) \in A \times [P]\}$.
We model these event sequences using a mixture of Hawkes processes. Specifically, for type $p$ event sequences between nodes $i$ and $j$ the intensity function at time $t$ is given by:
\begin{align}\label{mixture-network}
\lambda_{i,j,p}(t) = \mu_{z_i,z_j,p} + \sum_{l: t_l<t} f_{z_i,z_j}(t-t_l; \theta_{p_l,p})   
\end{align}
where $\mu_{z_i,z_j,p}$ is the base intensity for the event type $p$ corresponding to the nodes $i,j$ and $z_i \in [K], z_j \in [K]$ denote their cluster memberships. 
Notice that in the scenario where the nodes do not interact with each other and events correspond only to the individual node, the problem can be reduced to our setting as follows. It is a special case of our formulation by considering a bipartite network with the same set of nodes on each side with links mapping to the different event types between the corresponding duplicate nodes.

\subsection{ELBO}
The log-likelihood of our sequence when the labels of the nodes are known is given by 
$$\log \big(\sum_{k_1,k_2} \pi_{k_1,k_2} \mathrm{HP}^{\delta}(\vs|\bm \mu_{k_1,k_2},\bm \theta_{k_1,k_2}) \big).$$ due to summation inside the logarithm. 
Instead, we work with the evidence lower bound (ELBO) defined as 
\begin{align}
    \mathrm{ELBO}(\bm \Theta)= \E_{q(\bm z)}[L^{\delta}(\bm \Theta,\bm z)]- \E_{q(\bm z)}[\log q(\bm z)]
\end{align}
for a properly chosen distribution $q(\bm z)$, where $L^{\delta}(\bm \Theta, \bm z)$ is the complete log-likelihood defined as 
%
\begin{align}
    &L^{\delta}(\bm \Theta, \mZ) \triangleq \sum_{(i,j) \in A}\sum_{k_1=1}^K\sum_{k_2=1}^K z_{i,j,k_1,k_2}\Big[ \log \mathrm{HP}^{\delta}(\vs_{i,j}|\bm \mu_{k_1,k_2},\bm \theta_{k_1,k_2}) + \log \pi_{k_1,k_2} \Big]\nonumber\\
    %
    %
    &=  \sum\limits_{\tau=1}^{T/\delta}\sum_{(i,j) \in A} \sum\limits_{p=1}^{P}\sum_{k_1=1}^K\sum_{k_2=1}^K z_{i,j,k_1,k_2}\big[ x^{\tau}_{i,j,p} \log (\lambda_{i,j,k_1,k_2,p}(\tau\delta)\big)  \big] \nonumber\\
    &\quad -\sum\limits_{\tau=1}^{T/\delta}\sum_{(i,j) \in A} \sum\limits_{p=1}^{P}\sum_{k_1=1}^K\sum_{k_2=1}^K z_{i,j,k_1,k_2}\big[ \delta   \lambda_{i,j,k_1,k_2,p}(\tau\delta) \big]  + \sum_{(i,j) \in A} \sum_{k_1=1}^K\sum_{k_2=1}^K z_{i,j,k_1,k_2} \log \pi_{k_1,k_2}  \nonumber\\
    &=  \sum\limits_{\tau=1}^{T/\delta} L_{\tau}^{\delta}(\bm \Theta, \bm z)  + \sum_{(i,j) \in A} \sum_{k_1=1}^K\sum_{k_2=1}^K z_{i,j,k_1,k_2} \log \pi_{k_1,k_2} 
\end{align}
is the complete log-likelihood function.
The additive nature of $L_{\tau}^{\delta}(\bm \Theta, \bm z)$ allows for adopting online algorithms to maximize the ELBO function. We assume that $q(\bm z)$ takes the simple form $q(\bm z)=\prod_{n=1}^N q(\vz_n)$ where $q(\vz_n)\sim \mathrm{multinom}(\bm\alpha_n)$ and $\vz_n\in [K]$. Therefore,
\begin{align}
    \mathrm{ELBO}(\bm \Theta)&= \sum\limits_{\tau=1}^{T/\delta} \E_{q(\bm z)}L_{\tau}^{\delta}(\bm \Theta, \bm z)  + \sum_{n=1}^N \sum_{k_1=1}^K\sum_{k_2=1}^K \alpha_{i,k_1}\alpha_{j,k_2} \log \pi_{k_1,k_2}  \nonumber\\
    &\qquad- \sum_{(i,j)\in A} \sum_{k_1=1}^K\sum_{k_2=1}^K\alpha_{i,k_1}\alpha_{j,k_2} \log \alpha_{i,k_1}\alpha_{j,k_2}
    %
\end{align}
with
\begin{align}
 &\E_{q(\bm z)}L_{\tau}^{\delta}(\bm \Theta, \bm z)  = \sum_{(i,j)\in A} \sum\limits_{p=1}^{P} \sum_{k_1=1}^K\sum_{k_2=1}^K \alpha_{i,k_1} \alpha_{j,k_2}\big[ x^{\tau}_{i,j,p} \log (\lambda_{k_1,k_2,p}(\tau\delta)\big)  \big] \nonumber\\
&\qquad\qquad-\sum_{(i,j)\in A} \sum\limits_{p=1}^{P} \sum_{k_1=1}^K\sum_{k_2=1}^K \alpha_{i,k_1} \alpha_{j,k_2}\big[ \delta   \lambda_{k_1,k_2,p}(\tau\delta) \big] .
 \end{align}
The OMMHP can be applied here in a similar fashion to Section \ref{sec:algo}. 